\documentclass{article}
\usepackage{arxiv}
\usepackage[utf8]{inputenc}
\usepackage[T1]{fontenc}
\usepackage{url}
\usepackage{booktabs}
\usepackage{amsfonts}
\usepackage{nicefrac}
\usepackage{microtype}
\usepackage{graphicx}
\usepackage{doi}
\usepackage[svgnames]{xcolor}

\definecolor{ind_navyblue}{rgb}{0,0,0.50}

\hypersetup{colorlinks,breaklinks,
         urlcolor=ind_navyblue,
         linkcolor=ind_navyblue,
         citecolor=ind_navyblue}

\title{A perspective on fluid mechanical environments for challenges in reinforcement learning} 

  \author{
  Shruti Mishra$^1$ \hspace{0.5em}
  Michael Chang$^2$ \hspace{0.5em}
  Vamsi Spandan$^1$ \hspace{0.5em}
  Shmuel M.~Rubinstein$^3$
  }
  \affiliations{$^1$Sony AI \hspace{2.5em} $^2$Cohere Labs Community \hspace{2.5em} $^3$The Hebrew
  University of Jerusalem}
  \emails{}
  \setrunningtitle{A perspective on fluid mechanical environments for challenges in reinforcement learning}

\date{July 1, 2025}	

\hypersetup{
pdftitle={A perspective on fluid mechanical environments for challenges in reinforcement learning},
pdfsubject={CS.LG},
pdfauthor={S.~Mishra, M.~Chang, V.~Spandan, S.M.~Rubinstein},
pdfkeywords={reinforcement learning, nonstationarity, fluid mechanics, nonlinear dynamics},
}

\begin{document}
\begingroup
\renewcommand\thefootnote{}
\footnotetext{Author for correspondence: Shruti Mishra; email: \texttt{shrutimishraphd@gmail.com}}
\endgroup

\renewcommand{\sectionautorefname}{Section}
\renewcommand{\subsectionautorefname}{Section}
\renewcommand{\subsubsectionautorefname}{Section}

\maketitle
\section*{Abstract}
We consider the challenge of developing agents that efficiently interact with high-dimensional, evolving environments, towards a view of practical reinforcement learning (RL) agents interacting with open worlds, of which they witness and affect only a small part. We argue that canonical fluid mechanics problems, and their simulations, present a compelling testbed for the development of such methods. These problems arise in nonlinear instabilities, where small disturbances can grow to transform the dynamics of a system. Nonlinear instabilities represent several open scientific challenges with industrial applications --- the droplet breakup of a liquid jet, mixing at an interface between two fluids, and the appearance of unusually tall rogue waves in the ocean. In these settings, agents may leverage preserved representations across the changing dynamics to learn efficiently. 

We present two problem descriptions of agents interacting with a fluid mechanical environment, and describe the state and action spaces, and reward functions, for these agents. For these examples, we specify the aspects of the environment which are nonstationary and the preserved invariances. 
We note Dedalus and JAX-CFD as open-source simulators that can be used for the development of reinforcement learning methods~\citep{burns2016dedalus, kochkov2021machine}. We demonstrate the use of Dedalus for environment generation by creating RL agents that learn to navigate in a stationary environment that is simulated using Dedalus. This sets the stage for future development of RL agents that learn to meaningfully interact with simulated environments that represent scientific challenges in natural and industrial flows.

\section{Introduction}
\label{sec:intro}

Over the last decade, artificial environments such as games have catalyzed the development of some of the most advanced artificial methods for reinforcement learning~\citep{mnih2015human, silver2017mastering, silver2018general, vinyals2019grandmaster, wurman2022outracing, perolat2022mastering}. 
In advancing reinforcement learning methods towards real-world settings that include the understanding of natural phenomena, a number of challenges remain~\citep{dulac2021challenges}.
A key challenge is that of agents efficiently interacting with an evolving, high-dimensional, environment, beyond the paradigm of statistically stationary environment interactions, towards the setting of a finite agent learning from interactions in a big world~\citep{javed2024big}.
Intelligent agents operating in the real world navigate changing environmental dynamics. To do so, agents can leverage knowledge of how the environment is structured to inform meaningful behavior. 
For instance, animals in urban environments adapt their behaviors to respond to the changes created by human activity --- birds have changed the pitch of their song, reducing disruption by the sounds of an urban environment (e.g.~\citet{slabbekoorn2006cities}), and squirrels have learned to seek ways to obtain food from humans. 

In nonlinear instabilities in fluid flows, small disturbances can grow and fundamentally alter the dynamics of a system --- 
a small ripple on a water surface can grow into an unusually large rogue wave and a small fluctuation in an engine can trigger destructive oscillations. The growth of small disturbances to a flow can transform pre-existing flow patterns. 
When disturbances to a flow evolve, flows can transition towards turbulence, a class of canonical multiscale phenomena.
With a wide variety of fundamental scientific questions and relevance for industrial applications, hydrodynamic instabilities have naturally been associated with deep scientific activity~\citep{chandrasekhar2013hydrodynamic}.
However, nonlinear instabilities are associated with yet unsolved scientific mysteries such as a discrepancy between computational predictions and experimental observations on the growth of a mixing layer at an interface between two fluids. Industrial applications include predicting thermoacoustic oscillations in combustors for energy production and heat flow during boiling in power plants. 

Across the dynamics associated with small waves and destructive rogue waves in an ocean, even as energy is transferred across the spatial scales, mathematical representations associated with conservation of mass, momentum and energy are preserved to a good approximation. Agents operating in highly changing fluid flow environments can thus leverage such preserved representations while acting meaningfully across changes in flow structures. 
Specifically, fluid flows are well modeled by the Navier--Stokes equations. These equations represent conservation of mass and conservation of momentum. With such known parsimonious models, flows have representations that are preserved across critical transitions that are characteristic of instabilities in fluid flows~\citep{chandrasekhar2013hydrodynamic}. 
With known and learnable representations that can be leveraged, flows with nonlinear instabilities can form a compelling testbed for the development and demonstration of reinforcement learning agents efficiently interacting with evolving, high-dimensional, environments. 

Objectives for reinforcement learning agents in flows with nonlinear instabilities can include predicting, meaningfully affecting, and controlling flows with apparently changing dynamics. The proposed paradigm of reinforcement learning in flows with nonlinear instabilities is one of an interplay of apparently changing dynamics and some preserved representations. 

\section{Stationarity in representation: Invariances in fluid flows}
\label{sec:stationarity-examples}

Physical systems are studied using theoretical, computational and experimental approaches. This section presents some representations in fluid flows that remain preserved, even across transient dynamics in the observable field variables such as velocity. The text, ``Fluid Mechanics'' \citep{kundu2002fluid}, provides a mathematical introduction to the subject, and ``Elementary Fluid Mechanics'' \citep{acheson1990elementary} presents a more concise description. 

When represented in terms of field variables such as velocity, pressure and density, varying over spatial domain, and evolving over time, fluid flows are naturally high-dimensional. As an example, flows can require thousands of grid points over even millimetre-sized spatial domains with uniform resolution, evolving over milliseconds, for adequate numerical resolution e.g.~\citep{mishra2022computing}. At the same time, such flows are represented mathematically by a small set of governing equations which determine the evolution of the corresponding flows. 

Mathematical models are a powerful tool for capturing relevant physics while abstracting processes that happen at resolutions different from the problem in consideration. Fluid mechanics itself is an example of such an abstraction; interactions and collisions at a molecular level are coarse-grained and expressed in terms of equations for the conservation of mass, momentum and energy for a continuum -- the fluid. 

The Navier--Stokes equations, indicated for incompressible flows by \autoref{eqn:navier-stokes}, 
\begin{subequations}
\label{eqn:navier-stokes}
\begin{align}
&\boldsymbol{\nabla}\cdot{\boldsymbol{u}}=0,
&&\text{conservation of mass}, 
\label{eqn:cont} \\
&\boldsymbol{u}_{t} 
+ \left(\boldsymbol{u} \cdot \boldsymbol{\nabla}\right)\boldsymbol{u} =
-\frac{\boldsymbol{\nabla}p}{\rho} 
+ \nu \nabla^2 \boldsymbol{u},
&&\text{conservation of momentum}, 
\label{eqn:mom}
\end{align}
\end{subequations}
are a powerful representation of fluid mechanical systems that represent the conservation of mass and momentum, where $\boldsymbol{u}$ is the fluid velocity, $p$ is pressure and $\rho$ is density. An ubiquitous abstraction within the framework of the Navier--Stokes equations is the quantity \emph{viscosity}, often expressed as a constant for Newtonian fluids i.e.~$\nu$ in \autoref{eqn:mom}. Viscosity represents the macroscopic effect of mixing through movement of particles at a molecular level in terms of the smoothing of gradients of flow-field variables such as velocity. Across several flows with transient flow structures, such as the rapidly changing flow structures around a racecar, the Navier--Stokes equations with a constant viscosity of air remain a good approximation of the flow dynamics. 

Abstractions of various types have been used to further enable the solution of these equations for particular problems. For instance, using knowledge of the relevant physical processes to remove terms that are deemed to be small compared to other terms, is done to enable tractability of the problem. These conservation relationships can remain preserved across the relevant dynamics, and arguably provide a world model for intelligent agents. 

Computational methods to solve the Navier--Stokes equations rely on further abstractions of flow-field quantities. For instance, finite-difference methods assume that the continuous quantities can be represented by discretized counterparts that vary on a grid. Spectral methods represent the underlying field variables in terms of periodic functions over a domain of interest. Finite-difference methods and spectral methods are two broad categories of a rich set of scientifically validated representations of the Navier--Stokes equations. 

\section{Nonstationary environments with stationary representations}
Having noted a variety of stationary representations that arise in fluid flows in \autoref{sec:stationarity-examples}, this section describes examples of reinforcement learning agents interacting with nonstationary fluid environments which also have known stationary representations. 

\subsection{An agent in a transitioning pipe flow}
\begin{figure}[h]
    \centering
    \includegraphics[width=0.8\linewidth]{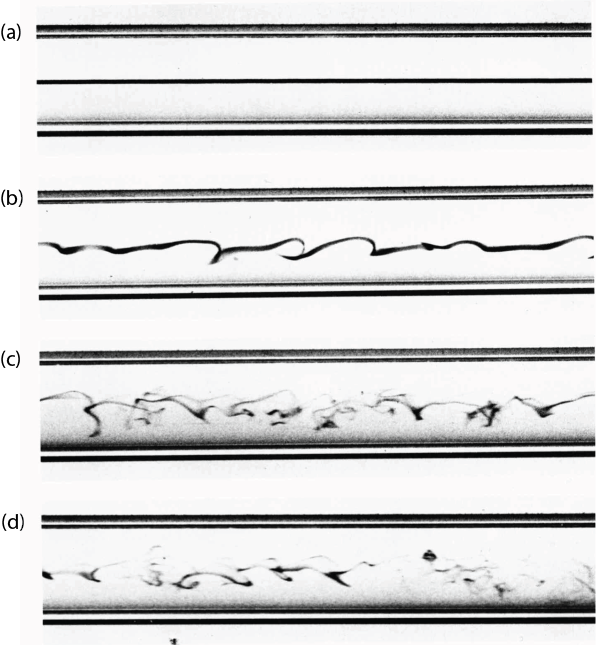}
    \caption{
    Four snapshots, from top to bottom, illustrating the evolution of a filament of colored water in a pipe flow~\citep{van1982album}. In all panels, the colored water is indicated in the darker color. (a) The colored water is undisturbed, and the filament occupies a region in the middle of the pipe. (b--d) As the speed of the flow is increased, the filament becomes increasingly unstable, and transitions towards a disordered state. The transition towards such disordered states occurs via the nonlinear interactions of disturbances with each other and with the background flow in the pipe. The book, ``An Album of Fluid Motion'' by \citet{van1982album} is a pictographic guide to the diversity of flow patterns observed in nature and in engineering applications.}
    \label{fig:reynolds-pipe}
\end{figure}
This section describes the setup of a reinforcement learning agent interacting with a flow in a pipe that transitions towards turbulence through the growth and nonlinear interactions of perturbations to a mean flow profile. Turbulence is a canonical multiscale phenomenon; interactions in a turbulent flow range from the largest length scale in the flow, dominated by inertial interactions, to the smallest scale, dominated by viscous interactions, e.g.~\citep{mckeown2020turbulence}. An agent that takes the form of an active particle in a turbulent flow, buffeted by flow structures, experiences interactions with flow structures across these length scales, typically spanning multiple orders of magnitude. Its world is big~\citep{javed2024big}. 

The flow of water in a pipe is well described by the Navier--Stokes equations for conservation of mass and momentum (\autoref{eqn:navier-stokes}). The geometry of a three-dimensional pipe flow comprises a circular cross section of radius $R$, uniform along the length of the pipe, and the flow is driven by a pressure gradient along the length of the pipe. With boundary conditions $\boldsymbol{u} = \boldsymbol{0}$ at the pipe walls, $r=R$, the mean flow profile for a fully developed pipe flow, downstream of the inlet, is 
\begin{equation}
\label{eqn:poiseuille-flow}
    \boldsymbol{u} = u \boldsymbol{e}_z = U_0 \left(1-(r/R)^2\right) \boldsymbol{e}_z,
\end{equation}
where $\boldsymbol{e}_z$ is the unit vector along the length of the pipe and $U_0$ is a constant. 

Pipe flow is a canonical flow that is stable to infinitesimal disturbances, but can transition to turbulence via the growth and nonlinear interactions of disturbances to the parabolic mean flow profile~\citep{reynolds1883xxix}. This is illustrated in \autoref{fig:reynolds-pipe}.
In this environment, the observations for the agent are a three-dimensional representation for the velocity field and the pressure field; $\mathcal{S} \equiv [\boldsymbol{u},p]$, over a distance between $z=0$ and $z=L$. The action chosen by the agent is equivalent to injecting and moving obstacles in the flow that disturb the uniform cross section of the flow. For simplicity, the agent can move the obstacle at a fixed speed $|\boldsymbol{u}_\mathrm{obstacle}|$ and can change the direction of movement of the obstacle at a rate of $\dot \theta$, chosen by the agent i.e.~$\mathcal{A} \equiv \dot\theta$. At the locations $\boldsymbol{x}$ occupied by the obstacle, the flow velocity $\boldsymbol{u} = \boldsymbol{0}$. This creates a perturbation to the mean profile specified in \autoref{eqn:poiseuille-flow}. The obstacle can be specified as a small, rigid sphere of diameter $\delta < R$. 
The agent's goal is to create the maximum possible disturbance to the mean profile. Accordingly, the reward $r$ can be $\|\boldsymbol{u}-U_0 \left(1-(r/R)^2\right) \boldsymbol{e}_z\|_2$ in a specified time horizon $T$.
Through being incentivized to create disturbances to the flow, the RL agent can create an environment that is highly nonstationary at the level of the state description $\mathcal{S}$, e.g.~\citep{avila2011onset}. Even as the agent can aim to do so, the description of the flow remains invariant at the level of the governing Navier--Stokes equations \eqref{eqn:navier-stokes}.

\subsection{An agent in an evolving cellular flow}
\label{sec:rl-flow-cellular}
\begin{figure}[h]
    \centering
    \includegraphics[width=0.8\linewidth]{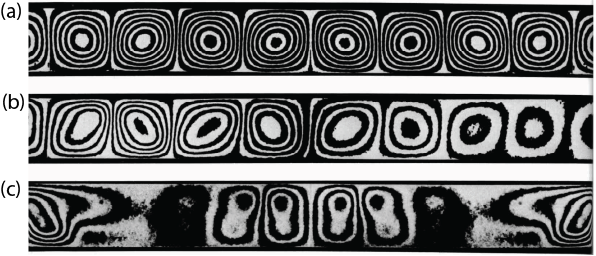}
    \caption{Three snapshots, (a--c), of fluid flow between two plates~\citep{van1982album}. In each panel, the bottom plate is heated. Under small disturbances to a static fluid between the plates, a flow evolves into convection rolls. In panel (a) the plates are maintained at fixed temperatures. In panel (b), there is a temperature gradient in the temperature of the bottom plate, along the length of the plate. In panel (c), the plate and fluid is subject to an external rotational motion.}
    \label{fig:convection-rolls}
\end{figure}
Reinforcement learning studies have considered particle swimmers in steady fluid flows i.e.~with velocity and pressure constant in time, with a limited state space~\citep{colabrese2017flow, gustavsson2017finding}. The strength of the background flow field affects the optimal strategy for a swimmer~\citep{colabrese2017flow}. 
The setup of a reinforcement learning agent in a nonstationary environment can consist of a particle in an evolving cellular flow, such as that illustrated in \autoref{fig:convection-rolls}. An example setup for a two dimensional flow is described by 
\begin{equation}
    \label{eqn:vortex-decay}
    \omega = -2U_0 \cos(x) \cos(y) \exp\left(-2\nu t\right),
\end{equation}
where $\omega = \boldsymbol{\nabla} \times \boldsymbol{u}$. This flow is a solution to \autoref{eqn:navier-stokes} for the initial condition $\omega(x, y, t=0) \equiv -2U_0 \cos(x) \cos(y)$ in an infinite two-dimensional space. In this flow, the state space $\mathcal{S}\equiv [\omega, \theta]$, where $\theta$ is the orientation of a particle in the flow. The action space $\mathcal{A} \equiv \dot{\theta} $ is the change in orientation of the particle. The reward $R \equiv y(s') - y(s)$ defines the change in position between subsequent states, in the preferred direction of movement along $y$. This setup is similar to that of \citet{colabrese2017flow}, who used reinforcement learning to train particle swimmers moving through a stationary flow pattern described by $\omega = -2U_0 \cos(x) \cos(y)$. 

\begin{figure}
    \centering
    \includegraphics[width=0.7\linewidth]{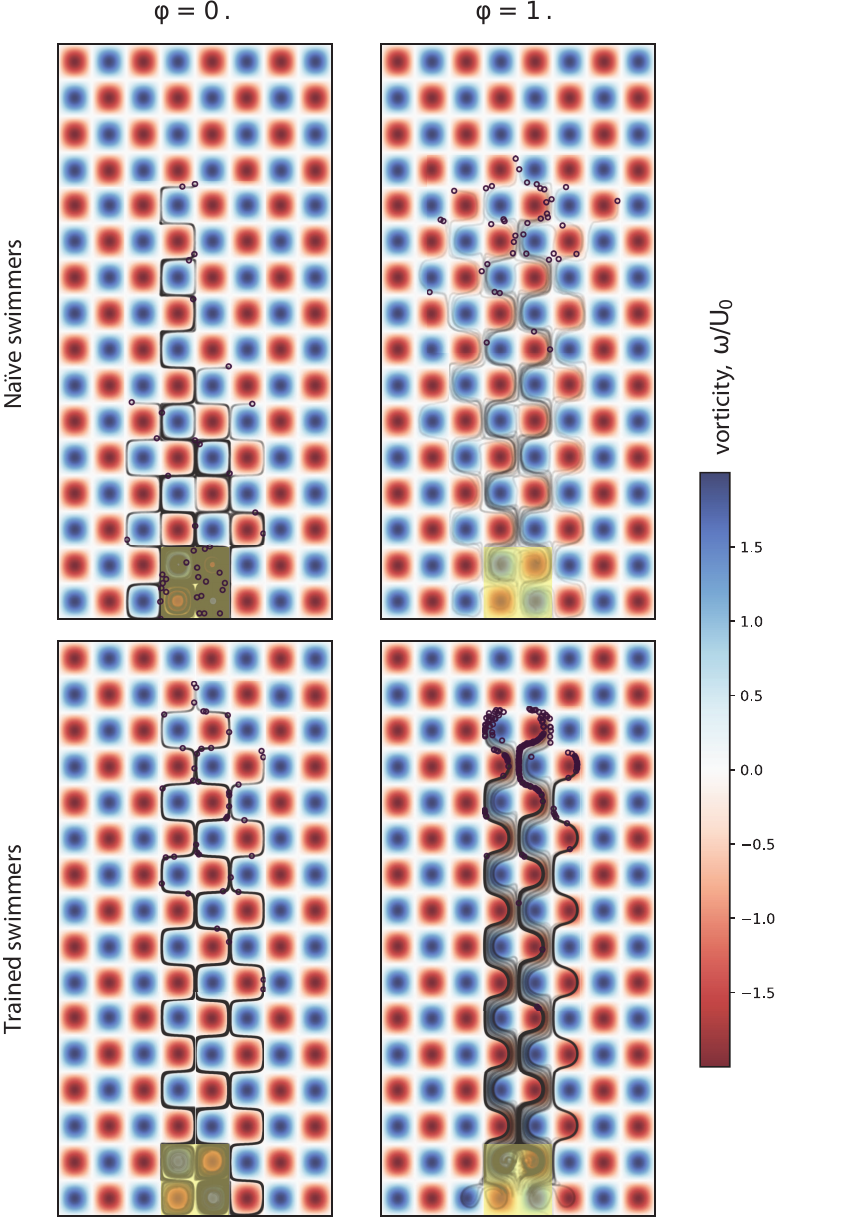}
    \caption{Swimmers navigating through a steady (unchanging) vortex field. The strength of the vorticity field is shown in the red--blue color scheme. The left column shows swimmers that are weaker than the surrounding flow, and thus have to learn to navigate the structure of the background flow. The right column illustrates swimmers that have a strength comparable to that of the background flow. The grey lines in the panels indicate the trajectories of 250 swimmers, initialized in the region indicated by the yellow square. The dark purple circles indicate final positions attained by the swimmers in a fixed time duration. The top row shows naïve swimmers which always attempt to orient themselves upwards. The bottom row shows swimmers trained using tabular Q-learning. This demonstrates the training setup of \citet{colabrese2017flow}, and shows strategies (right panel) for navigating steady flows of different strengths. We set up the flow using Dedalus~\citep{burns2016dedalus}. Simulation and training parameters are specified in \autoref{tbl:app}.}
    \label{fig:rl-microswimmer}
\end{figure}

The time dependence in the flow described by \autoref{eqn:vortex-decay}, through the term $e^{-\nu t}$, means that the flow, and thus the environment experienced by a particle swimmer through the flow, is inherently nonstationary.
In this flow, preserved invariances include the governing Navier--Stokes equations \eqref{eqn:navier-stokes}, a periodic structure in the flow, and the material property viscosity. As indicated in \autoref{fig:rl-microswimmer}, the learned strategy for a swimmer navigating through this structure changes as the strength of a stationary flow relative to the swimmer strength changes. As such, a swimmer navigating an evolving flow can leverage the evolving structure to determine new policies for traversal. 

\subsection{Flow control in the literature}
Flow control using RL has been considered in prior literature, largely towards achieving a statistically stationary regime. \citet{beintema2020controlling} consider flow control in Rayleigh--B{\'e}nard convection, a classical system for studying thermal convection, a model system for cloud dynamics~\citep{kundu2002fluid,drazin2004hydrodynamic,bodenschatz2000recent}
and the chosen system for the development of several supervised learning algorithms. 
In the Kuramoto--Sivashinsky equations, a classical system for spatiotemporal chaos, \citet{weissenbacher2025reinforcement} consider flow control towards mitigating chaos, and creating a statistically stationary regime. 

\section{Simulators for reinforcement learning in fluid flows}
Virtual environments created by simulators have enabled the advancement and demonstration of reinforcement learning algorithms~\citep{barth2018distributed, haarnoja2018soft, hafner2019learning}. Examples include the MuJoCo physics engine and Isaac Gym, which form testbeds for bio-inspired and robotics domains~\citep{todorov2012mujoco, makoviychuk2021isaac}. In order for flows with nonlinear instabilities to be adopted as an interesting domain for advancing reinforcement learning methods, the adoption of a relevant simulator may hold similar significance. This section presents two simulators that may be used for the development of reinforcement learning algorithms in evolving fluid flows, followed by a potential direction for the development of simulators.

\subsection{Open-source simulators: Dedalus and JAX-CFD}
Dedalus is a general-purpose simulator for differential equations~\citep{burns2016dedalus}. Analogous to MuJoCo leveraging approximations to allow for simulations to be practically feasible, the number of spectral coefficients in Dedalus can be chosen for speed versus fidelity. Dedalus has a pre-existing set of examples of flows with nonlinear instabilities. In machine learning, Dedalus has been used as the simulator for super-resolution in Rayleigh--B\'enard convection, which is a model system for cloud dynamics~\citep{jiang2020meshfreeflownet}. 

As an example, we replicate the environment of \citet{colabrese2017flow}, and demonstrate the use of Dedalus to create the environment of a cellular flow pattern described in \autoref{sec:rl-flow-cellular}, and illustrated in \autoref{fig:convection-rolls}. Briefly, an agent is in a flow described by a stationary vorticity field,
\begin{equation}
\label{eqn:vortex-stationary}
\omega = -2U_0 \cos(x) \cos(y).
\end{equation}
Similar to the setup of \citet{colabrese2017flow}, a swimmer in the two dimensional vorticity field has a fixed speed and can choose to change its direction vector. The goal of the agent is to traverse upwards, with the state and action spaces, and reward function specified in \autoref{sec:rl-flow-cellular}. For the observation, the vorticity is discretized into three values and the direction of the swimmer is discretized into four values, $\{\uparrow, \rightarrow, \leftarrow, \downarrow\}$. The action for the preferred swimming direction is similarly discretized into four values. \autoref{fig:rl-microswimmer} shows trajectories for swimmers for increasing amounts of training, illustrating results similar to those of \citet{colabrese2017flow}, with the environment set up in Dedalus. 

JAX-CFD is a GPU simulator for fluid mechanics, designed for coupling with machine learning methods~\citep{kochkov2021machine}. JAX-CFD has largely been used with supervised learning methods, and more recently for problems in flow control~\citep{alhashim2025control}. 

\subsection{Simulators based on learned surrogate models}
The speed of a simulator is a limiting factor for the development of reinforcement learning methods. Surrogate models have been an area of development in the machine learning community in recent years. These models aim to learn key physics during training time, to allow for relatively cheap inference compared to traditional simulations of fluid mechanics~\citep{fonda2019deep,jiang2020meshfreeflownet,pfaff2021learning,lusch2018deep,sanchez2020learning, wiewel2019latent,wu2020enforcing}. Surrogate models have the potential to provide for simulators that can allow for fast computation of the environment dynamics in the reinforcement learning setting. 

\section{Summary and future directions}
The perspective in this paper presents scientific challenges associated with nonlinear instabilities in fluid mechanics as a source of problem formulations for reinforcement learning algorithms, particularly for learning efficiently nonstationary environments. Nonlinear instabilities result in highly changing dynamics, with preserved invariances in the form of conservation equations and potentially in the form of material properties such as viscosity. These invariances can be leveraged by agents towards developing agents that can meaningfully interact with changing environments. We described two examples of nonlinear instabilities. Both the examples are modeled by the Navier--Stokes equations, a well established representation for fluid mechanical systems. For the instabilities, we specified the state space, action space and a reward function for a reinforcement learning agent interacting with the instabilities. 

Simulated environments have been instrumental in the development of artificial reinforcement learning methods.
We described two simulators, Dedalus and JAX-CFD, as avenues for creating environments for RL agents that interact with nonstationary environments. For a stationary environment related to the nonstationary cellular flow described in Section \ref{sec:rl-flow-cellular}, we reproduced results from the literature using Dedalus to create the environment. Relatedly, we noted surrogate models as a source of potential environments for RL agents in the future. 

The description of nonlinear instabilities with preserved invariances, together with a demonstration of Dedalus to create environments for RL agents, sets the stage for the integration of RL agents that can learn to leverage known invariances to act efficiently in evolving environments. 
Future directions can incorporate the environment capabilities of a simulator like Dedalus, to integrate it with nonstationary environments such as those created in nonlinear instabilities. While we have described instabilities in fluid flows, with remarkably preserved invariances in the form of the Navier--Stokes equations, as a source of problem descriptions for advancing reinforcement learning methods, instabilities exist beyond fluid flows. For example, there are instabilities in the dynamics of solid objects. Dedalus includes the capability to simulate some of these dynamics, such the bending of an elastic sheet. 

\section*{Acknowledgements}
S.M.~thanks Yannis Hardalupas and Miles Cranmer for discussions of fluid instabilities, Daniel Fortunato for a discussion on Dedalus, and Jordan Hoffmann for reading drafts of this paper. S.M.~acknowledges partial funding support from CQDM under the FACS/Acquité project and a G-Research Grant for Early Career Researchers. We thank the reviewer for feedback on an earlier version of this paper. 

\bibliography{references}

@book{acheson1990elementary,
  title={Elementary fluid dynamics},
  author={Acheson, David J},
  year={1990},
  publisher={Oxford University Press}
}

@article{alhashim2025control,
  title={Control of flow behavior in complex fluids using automatic differentiation},
  author={Alhashim, Mohammed G and Hausknecht, Kaylie and Brenner, Michael P},
  journal={Proceedings of the National Academy of Sciences},
  volume={122},
  number={8},
  pages={e2403644122},
  year={2025},
  publisher={National Academy of Sciences}
}

@article{avila2011onset,
  title={The onset of turbulence in pipe flow},
  author={Avila, Kerstin and Moxey, David and De Lozar, Alberto and Avila, Marc and Barkley, Dwight and Hof, Bj{\"o}rn},
  journal={Science},
  volume={333},
  number={6039},
  pages={192--196},
  year={2011},
  publisher={American Association for the Advancement of Science}
}

@inproceedings{barth2018distributed,
  title={Distributed Distributional Deterministic Policy Gradients},
  author={Barth-Maron, Gabriel and Hoffman, Matthew W and Budden, David and Dabney, Will and Horgan, Dan and Dhruva, TB and Muldal, Alistair and Heess, Nicolas and Lillicrap, Timothy},
  booktitle={International Conference on Learning Representations},
  year={2018}
}

@article{beintema2020controlling,
  title={Controlling {R}ayleigh--{B}{\'e}nard convection via reinforcement learning},
  author={Beintema, Gerben and Corbetta, Alessandro and Biferale, Luca and Toschi, Federico},
  journal={Journal of Turbulence},
  volume={21},
  number={9-10},
  pages={585--605},
  year={2020},
  publisher={Taylor \& Francis}
}

@article{bodenschatz2000recent,
  title={Recent developments in {R}ayleigh--{B}{\'e}nard convection},
  author={Bodenschatz, Eberhard and Pesch, Werner and Ahlers, Guenter},
  journal={Annual review of fluid mechanics},
  volume={32},
  number={1},
  pages={709--778},
  year={2000},
  publisher={Annual Reviews 4139 El Camino Way, PO Box 10139, Palo Alto, CA 94303-0139, USA}
}

@article{burns2016dedalus,
  title={Dedalus: Flexible framework for spectrally solving differential equations},
  author={Burns, Keaton J and Vasil, Geoffrey M and Oishi, Jeffrey S and Lecoanet, Daniel and Brown, Benjamin},
  journal={Astrophysics Source Code Library},
  pages={ascl--1603},
  year={2016}
}

@book{chandrasekhar2013hydrodynamic,
  title={Hydrodynamic and {H}ydromagnetic {S}tability},
  author={Chandrasekhar, Subrahmanyan},
  year={2013},
  publisher={Courier Corporation}
}

@article{colabrese2017flow,
  title={Flow navigation by smart microswimmers via reinforcement learning},
  author={Colabrese, Simona and Gustavsson, Kristian and Celani, Antonio and Biferale, Luca},
  journal={Physical review letters},
  volume={118},
  number={15},
  pages={158004},
  year={2017},
  publisher={APS}
}

@book{drazin2004hydrodynamic,
  title={Hydrodynamic stability},
  author={Drazin, Philip G and Reid, William Hill},
  year={2004},
  publisher={Cambridge university press},
  edition={2}
}

@article{dulac2021challenges,
  title={Challenges of real-world reinforcement learning: definitions, benchmarks and analysis},
  author={Dulac-Arnold, Gabriel and Levine, Nir and Mankowitz, Daniel J and Li, Jerry and Paduraru, Cosmin and Gowal, Sven and Hester, Todd},
  journal={Machine Learning},
  volume={110},
  number={9},
  pages={2419--2468},
  year={2021},
  publisher={Springer}
}

@article{fonda2019deep,
  title={Deep learning in turbulent convection networks},
  author={Fonda, Enrico and Pandey, Ambrish and Schumacher, J{\"o}rg and Sreenivasan, Katepalli R},
  journal={Proceedings of the National Academy of Sciences},
  volume={116},
  number={18},
  pages={8667--8672},
  year={2019},
  publisher={National Acad Sciences}
}

@article{gustavsson2017finding,
  title={Finding efficient swimming strategies in a three-dimensional chaotic flow by reinforcement learning},
  author={Gustavsson, Kristian and Biferale, Luca and Celani, Antonio and Colabrese, Simona},
  journal={The European Physical Journal E},
  volume={40},
  pages={1--6},
  year={2017},
  publisher={Springer}
}

@inproceedings{haarnoja2018soft,
  title={Soft actor-critic: Off-policy maximum entropy deep reinforcement learning with a stochastic actor},
  author={Haarnoja, Tuomas and Zhou, Aurick and Abbeel, Pieter and Levine, Sergey},
  booktitle={International conference on machine learning},
  pages={1861--1870},
  year={2018},
  organization={Pmlr}
}

@inproceedings{hafner2019learning,
  title={Learning latent dynamics for planning from pixels},
  author={Hafner, Danijar and Lillicrap, Timothy and Fischer, Ian and Villegas, Ruben and Ha, David and Lee, Honglak and Davidson, James},
  booktitle={International conference on machine learning},
  pages={2555--2565},
  year={2019},
  organization={PMLR}
}

@inproceedings{javed2024big,
  title={The big world hypothesis and its ramifications for artificial intelligence},
  author={Javed, Khurram and Sutton, Richard S},
  booktitle={Finding the Frame: An RLC Workshop for Examining Conceptual Frameworks},
  year={2024}
}

@inproceedings{jiang2020meshfreeflownet,
  title={{MeshfreeFlowNet}: a physics-constrained deep continuous space-time super-resolution framework},
  author={Jiang, Chiyu ``Max'' and Esmaeilzadeh, Soheil and Azizzadenesheli, Kamyar and Kashinath, Karthik and Mustafa, Mustafa and Tchelepi, Hamdi A and Marcus, Philip and Prabhat and Anandkumar, Anima and others},
  booktitle={SC20: International Conference for High Performance Computing, Networking, Storage and Analysis},
  pages={1--15},
  year={2020},
  organization={IEEE}
}

@article{kochkov2021machine,
  title={Machine learning--accelerated computational fluid dynamics},
  author={Kochkov, Dmitrii and Smith, Jamie A and Alieva, Ayya and Wang, Qing and Brenner, Michael P and Hoyer, Stephan},
  journal={Proceedings of the National Academy of Sciences},
  volume={118},
  number={21},
  pages={e2101784118},
  year={2021},
  publisher={National Academy of Sciences}
}

@book{kundu2002fluid,
  title={Fluid {M}echanics},
  author={Kundu, Pijush K and Cohen, Ira M and Dowling, David R},
  year={2012},
  publisher={Elsevier},
  isbn ={978-0-12-382100-3},
  edition={5}
}

@article{lusch2018deep,
  title={Deep learning for universal linear embeddings of nonlinear dynamics},
  author={Lusch, Bethany and Kutz, J Nathan and Brunton, Steven L},
  journal={Nature communications},
  volume={9},
  number={1},
  pages={1--10},
  year={2018},
  publisher={Nature Publishing Group}
}

@article{makoviychuk2021isaac,
  title={{Isaac Gym: High performance GPU-based physics simulation for robot learning}},
  author={Makoviychuk, Viktor and Wawrzyniak, Lukasz and Guo, Yunrong and Lu, Michelle and Storey, Kier and Macklin, Miles and Hoeller, David and Rudin, Nikita and Allshire, Arthur and Handa, Ankur and Slate, Gavriel},
  journal={arXiv preprint arXiv:2108.10470},
  year={2021}
}

@article{mckeown2020turbulence,
  title={Turbulence generation through an iterative cascade of the elliptical instability},
  author={McKeown, Ryan and Ostilla-M{\'o}nico, Rodolfo and Pumir, Alain and Brenner, Michael P and Rubinstein, Shmuel M},
  journal={Science advances},
  volume={6},
  number={9},
  pages={eaaz2717},
  year={2020},
  publisher={American Association for the Advancement of Science}
}

@article{mishra2022computing,
  title={Computing the viscous effect in early-time drop impact dynamics},
  author={Mishra, Shruti and Rubinstein, Shmuel M and Rycroft, Chris H},
  journal={Journal of Fluid Mechanics},
  volume={945},
  pages={A13},
  year={2022},
  publisher={Cambridge University Press}
}

@article{mnih2015human,
  title={Human-level control through deep reinforcement learning},
  author={Mnih, Volodymyr and Kavukcuoglu, Koray and Silver, David and Rusu, Andrei A. and Veness, Joel and Bellemare, Marc G. and Graves, Alex and Riedmiller, Martin and Fidjeland, Andreas K. and Ostrovski, Georg and Petersen, Stig and Beattie, Charles and Sadik, Amir and Antonoglou, Ioannis and King, Helen and Kumaran, Dharshan and Wierstra, Daan and Legg, Shane and Hassabis, Demis},
  journal={Nature},
  volume={518},
  number={7540},
  pages={529--533},
  year={2015},
  publisher={Nature Publishing Group}
}

@article{perolat2022mastering,
  title={Mastering the game of {Stratego} with model-free multiagent reinforcement learning},
 author={Perolat, Julien and de Vylder, Bart and Hennes, Daniel and Tarassov, Eugene and Strub, Florian and de Boer, Vincent and Muller, Paul and Connor, Jerome T. and Burch, Neil and Anthony, Thomas and McAleer, Stephen and Elie, Romuald and Cen, Sarah H. and Wang, Zhe and Gruslys, Audrunas and Malysheva, Aleksandra and Khan, Mina and Ozair, Sherjil and Timbers, Finbarr and Pohlen, Toby and Eccles, Tom and Rowland, Mark and Lanctot, Marc and Lespiau, Jean-Baptiste and Piot, Bilal and Omidshafiei, Shayegan and Lockhart, Edward and Sifre, Laurent and Beauguerlange, Nathalie and Munos, Remi and Silver, David and Singh, Satinder and Hassabis, Demis and Tuyls, Karl},
  journal={Science},
  volume={378},
  number={6623},
  pages={990--996},
  year={2022},
  publisher={American Association for the Advancement of Science}
}

@inproceedings{pfaff2021learning,
  title={Learning mesh-based simulation with graph networks},
  author={Pfaff, Tobias and Fortunato, Meire and Sanchez-Gonzalez, Alvaro and Battaglia, Peter W},
  booktitle={International Conference on Learning Representations},
  year={2021},
}

@article{reynolds1883xxix,
  title={{XXIX. An experimental investigation of the circumstances which determine whether the motion of water shall be direct or sinuous, and of the law of resistance in parallel channels}},
  author={Reynolds, Osborne},
  journal={Philosophical Transactions of the Royal society of London},
  number={174},
  pages={935--982},
  year={1883},
  publisher={The Royal Society London}
}

@inproceedings{sanchez2020learning,
  title={Learning to simulate complex physics with graph networks},
  author={Sanchez-Gonzalez, Alvaro and Godwin, Jonathan and Pfaff, Tobias and Ying, Rex and Leskovec, Jure and Battaglia, Peter},
  booktitle={International Conference on Machine Learning},
  pages={8459--8468},
  year={2020},
  organization={PMLR}
}

@article{silver2018general,
  title={A general reinforcement learning algorithm that masters chess, shogi, and Go through self-play},
  author={Silver, David and Hubert, Thomas and Schrittwieser, Julian and Antonoglou, Ioannis and Lai, Matthew and Guez, Arthur and Lanctot, Marc and Sifre, Laurent and Kumaran, Dharshan and Graepel, Thore and Lillicrap, Timothy and Simonyan, Karen and Hassabis, Demis},
  journal={Science},
  volume={362},
  number={6419},
  pages={1140--1144},
  year={2018},
  publisher={American Association for the Advancement of Science}
}

@article{silver2017mastering,
  title={{Mastering the game of Go without human knowledge}},
  author={Silver, David and Schrittwieser, Julian and Simonyan, Karen and Antonoglou, Ioannis and Huang, Aja and Guez, Arthur and Hubert, Thomas and Baker, Lucas and Lai, Matthew and Bolton, Adrian and Chen, Yutian and Lillicrap, Timothy and Hui, Fan and Sifre, Laurent and van den Driessche, George and Graepel, Thore and Hassabis, Demis},
  journal={Nature},
  volume={550},
  number={7676},
  pages={354--359},
  year={2017},
  publisher={Nature Publishing Group UK London}
}

@article{slabbekoorn2006cities,
  title={Cities change the songs of birds},
  author={Slabbekoorn, Hans and den Boer-Visser, Ardie},
  journal={Current biology},
  volume={16},
  number={23},
  pages={2326--2331},
  year={2006},
  publisher={Elsevier}
}

@inproceedings{todorov2012mujoco,
  title={{MuJoCo}: A physics engine for model-based control},
  author={Todorov, Emanuel and Erez, Tom and Tassa, Yuval},
  booktitle={2012 IEEE/RSJ international conference on intelligent robots and systems},
  pages={5026--5033},
  year={2012},
  organization={IEEE}
}

@book{van1982album,
  title={An album of fluid motion},
  author={Van Dyke, Milton},
  volume={176},
  year={1982},
  publisher={Parabolic Press Stanford}
}

@article{vinyals2019grandmaster,
  title={Grandmaster level in {StarCraft II} using multi-agent reinforcement learning},
  author={Vinyals, Oriol and Babuschkin, Igor and Czarnecki, Wojciech M and Mathieu, Micha{\"e}l and Dudzik, Andrew and Chung, Junyoung and Choi, David H and Powell, Richard and Ewalds, Timo and Georgiev, Petko and others},
  journal={nature},
  volume={575},
  number={7782},
  pages={350--354},
  year={2019},
  publisher={Nature Publishing Group}
}

@article{weissenbacher2025reinforcement,
  title={Reinforcement Learning of Chaotic Systems Control in Partially Observable Environments},
  author={Weissenbacher, Max and Borovykh, Anastasia and Rigas, Georgios},
  journal={Flow, Turbulence and Combustion},
  pages={1--22},
  year={2025},
  publisher={Springer}
}

@inproceedings{wiewel2019latent,
  title={Latent space physics: Towards learning the temporal evolution of fluid flow},
  author={Wiewel, Steffen and Becher, Moritz and Thuerey, Nils},
  booktitle={Computer graphics forum},
  volume={38},
  number={2},
  pages={71--82},
  year={2019},
  organization={Wiley Online Library}
}

@article{wu2020enforcing,
  title={Enforcing statistical constraints in generative adversarial networks for modeling chaotic dynamical systems},
  author={Wu, Jin-Long and Kashinath, Karthik and Albert, Adrian and Chirila, Dragos and Prabhat and Xiao, Heng},
  journal={Journal of Computational Physics},
  volume={406},
  pages={109209},
  year={2020},
  publisher={Elsevier}
}

@article{wurman2022outracing,
  title={Outracing champion {G}ran {T}urismo drivers with deep reinforcement learning},
  author={Wurman, Peter R and Barrett, Samuel and Kawamoto, Kenta and MacGlashan, James and Subramanian, Kaushik and Walsh, Thomas J and Capobianco, Roberto and Devlic, Alisa and Eckert, Franziska and Fuchs, Florian and Gilpin, Leilani and Khandelwal, Piyush and Kompella, Varun and Lin, HaoChih  and MacAlpine, Patrick and Oller, Declan and Seno, Takuma and Sherstan, Craig and Thomure, Michael D. and Aghabozorgi, Houmehr and Barrett, Leon and Douglas, Rory and Whitehead, Dion and Dürr, Peter and Stone, Peter and Spranger, Michael and Kitano, Hiroaki
  },
  journal={Nature},
  volume={602},
  number={7896},
  pages={223--228},
  year={2022},
  publisher={Nature Publishing Group UK London}
}
\bibliographystyle{agsm}

\newpage
\appendix
\section{Appendix: Parameters for simulation and training}
\begin{table}[h]
\caption{Parameter settings for simulation and training}
    \center
    \begin{tabular}{ll} 
    \toprule[1pt]
    \bf{Parameter} & \bf{Value(s)} \\ 
    \cmidrule[0.8pt]{1-2}
    \multicolumn{2}{l}{\textbf{\emph{Simulation parameters for Dedalus}}} \\ \cmidrule{1-2}
    Domain size, $L\equiv L_x \equiv L_y$ & $4 \pi$\\
    Resolution, $N \equiv N_x \equiv N_y$ & $512$ \\
    Number of vortices in simulation domain & $4$ \\
    Reference velocity, $U_0$ (\autoref{eqn:vortex-decay})& $1$\\ 
\cmidrule[0.8pt]{1-2}    \multicolumn{2}{l}{\textbf{\emph{Algorithm parameters for Q-learning}}} \\
    \cmidrule{1-2}
    Learning rate, $\alpha$ & $0.1$ \\
    Discount factor, $\gamma$ & $0.9$ \\
    Number of episodes & $1000$ \\
    Initial fraction of random actions, $\epsilon_{\mathrm{initial}}$ & $1$\\
    Final fraction of random actions at $700$ episodes, $\epsilon_{\mathrm{final}}$ & $1 \times 10^{-2}$ \\
    \bottomrule
    \end{tabular}
    \label{tbl:app}
\end{table}

\end{document}